# Coevolution Based Adaptive Monte Carlo Localization (CEAMCL)

Luo Ronghua & Hong Bingrong

Department of Computer Science, Harbin Institute of Technology, Harbin 150001, P. R. China,
ronghua75@yahoo.com.cn

*Abstract: An adaptive Monte Carlo localization algorithm based on coevolution mechanism of ecological species is proposed. Samples are clustered into species, each of which represents a hypothesis of the robot's pose. Since the coevolution between the species ensures that the multiple distinct hypotheses can be tracked stably, the problem of premature convergence when using MCL in highly symmetric environments can be solved. And the sample size can be adjusted adaptively over time according to the uncertainty of the robot's pose by using the population growth model. In addition, by using the crossover and mutation operators in evolutionary computation, intra-species evolution can drive the samples move towards the regions where the desired posterior density is large. So a small size of samples can represent the desired density well enough to make precise localization. The new algorithm is termed coevolution based adaptive Monte Carlo localization (CEAMCL). Experiments have been carried out to prove the efficiency of the new localization algorithm.*

## 1. Introduction

Self-localization, a basic problem in mobile robot systems, can be divided into two sub-problems: pose tracking and global localization. In pose tracking, the initial robot pose is known, and localization seeks to identify small, incremental errors in a robot's odometry (Leonard, J.J. & Durrant-Whyte, H.F., 1991). In global localization, however the robot is required to estimate its pose by local and incomplete observed information under the condition of uncertain initial pose. Global localization is a more challenging problem. Only most recently, several approaches based on probabilistic theory are proposed for global localization, including grid-based approaches (Burgard, W. *et al*, 1996), topological approaches (Kaelbling, L. P. *et al*, 1996) (Simmons, R. & Koenig, S., 1995), Monte Carlo localization (Dellaert, F. *et al*, 1999) and multi-hypothesis tracking (Jensfelt, P. & Kristensen, S., 2001) (Roumeliotis, S.I. & Bekey, G.A., 2000). By representing probability densities with sets of samples and using the sequential Monte Carlo importance sampling (Andrieu, C. & Doucet, A., 2002), Monte Carlo localization (MCL) can represent non-linear and non-Gaussian models well and focus the computational resources on regions with high likelihood. So MCL has attracted wide attention and has been applied in many real robot systems. But traditional MCL has some shortcomings. Since samples are actually drawn from a proposal density, if the observation density moves into one of the tails of the proposal density, most of the samples' non-normalized importance factors will be small. In this case, a large sample size is needed to represent the true posterior density to ensure stable and precise localization. Another problem is that samples often too quickly converge to a single, high likelihood pose. This might be undesirable in the case of localization in symmetric environments, where multiple distinct hypotheses have to be tracked for extended periods of time. How to get higher localization precision, to improve efficiency and to prevent premature convergence of MCL are the key concerns of the researchers. To make the samples represent the posterior density better, Thrun et al. proposed mixture-MCL (Thrun, S. *et al,* 2001), but it needs much additional computation in the sampling process. To improve the efficiency of MCL, methods adjusting sample size adaptively over time are proposed (Fox, D., 2003) (Koller, D. & Fratkina, R., 1998), but they increase the probability of premature convergence. Although clustered particle filters are applied to solve premature convergence (Milstein, A. *et al*, 2002), the method loses the advantage of focusing the computational resources on regions with high likelihood because it maintains the same sample size for all clusters.

In this paper, a new version of MCL is proposed to overcome those limitations. Samples are clustered into groups which are also called species. A coevolutionary model derived from competition of ecological species is introduced to make the species evolve cooperatively, so the premature convergence in highly symmetric environment can be prevented. The population growth model of species enables the sample size to be adjusted according to the total environment resources which represent uncertainty of the pose of the robot. And genetic operators are used for intra-species evolution to search for optimal samples in each species. So the samples can represent the desired posterior density better, and precise localization can be realized with a small size of sample. Compared with the traditional MCL, the new algorithm has the following advantages: (1) it can adaptively adjust the sample size during localization; (2) it can make stable localization in highly symmetric environment; (3) it can make precise localization with a small sample size.

## 2. Background

*2.1. Robot Localization Problem*

Robot localization is to estimate the current state $x_t$ of the robot, given the information about initial state and all the measurements $Y^t$ up to current time:

$$Y^t = \{y_t \mid t = 0,1,\cdots,t\} \quad (1)$$

Typically, the state $x_t$ is a three-dimensional vector including the position and direction of the robot, i.e. the pose of the robot. From a statistical point of view, the estimation of $x_t$ is an instance of Bayes filtering problem, which can be implemented by constructing the posterior density $p(x_t \mid Y^t)$. Assuming the environment is a Markov process, Bayes filters enable $p(x_t \mid Y^t)$ to be computed recursively in two steps.

Prediction step: Predicting the state of the next time-step with previous state $x_{t-1}$ according to the motion model $p(x_t \mid x_{t-1}, u_{t-1})$:

$$p(x_t \mid Y^{t-1})_{t-1} = \int p(x_t \mid x_{t-1}, u_{t-1}) p(x_{t-1} \mid Y^{t-1}) dx \quad (2)$$

Update step: Updating the state with the newly observed information $y_t$ according to the perceptual model $p(y_t \mid x_t)$:

$$p(x_t \mid Y^t) = \frac{p(y_t \mid x_t) p(x_t \mid Y^{t-1})}{p(y_t \mid Y^{t-1})} \quad (3)$$

*2.2. Monte Carlo localization (MCL)*

If the state space is continuous, as is the case in mobile robot localization, implementing equations (2) and (3) is not trivial. The key idea of MCL is to represent the posterior density $p(x_t \mid Y^t)$ by a set of weighted samples $S_t$:

$$S_t = \{(x_t^{(j)}, w_t^{(j)}) \mid j = 1,\cdots,N\} \quad (4)$$

Where $x_t^{(j)}$ is a possible state of the robot at current time $t$. The non-negative numerical factor $w_t^{(j)}$ called importance factor represents the probability that the state of robot is $x_t^{(j)}$ at time $t$. MCL includes the following three steps:

(1) Resampling: Resample $N$ samples randomly from $S_{t-1}$, according to the distribution defined by $w_{t-1}$;

(2) Importance sampling: sample state $x_t^{(j)}$ from $p(x_t \mid x_{t-1}^{(j)}, u_{t-1})$ for each of the $N$ possible state $x_{t-1}^{(j)}$; and evaluate the importance factor $w_t^{(j)} = p(y_t \mid x_t^{(j)})$.

(3) Summary: normalize the importance factors $w_t^{(j)} = w_t^{(j)} / \sum_{k=1}^{N} w_t^{(k)}$; and calculate the statistic property of sample set $S_t$ to estimate the pose of the robot.

*2.3. Coevolutionary algorithms*

Evolutionary algorithms (EAs), especially genetic algorithms, have been successfully used in different mobile robot applications, such as path planning (Chen, M. & Zalzala, A., 1995) (Potvin, J. *et al*, 1996), map building (Duckett, T., 2003) and even pose tracking for robot localization (Moreno, L. *et al*, 2002). But premature convergence is one of the main limitations when using evolutionary computation algorithms in more complex applications as in the case of global localization.

Coevolutionary algorithms (CEAs) are extensions of EAs. Based on the interaction between species, coevolution mechanism in CEAs can preserve diversity within the population of evolutionary algorithms to prevent premature convergence. According to the characters of interaction between species, CEAs can be divided into cooperative coevolutionary algorithms and competitive coevolutionary algorithms. Cooperative (also called symbiotic) coevolutionary algorithms involve a number of independently evolving species which together form complex structures. The fitness of an individual depends on its ability to collaborate with individuals from other species. Individuals are rewarded when they work well with other individuals and punished when they perform poorly together (Moriarty, D.E. & Miikkulainen, R., 1997) (Potter, M.A. & De Jong, K.A., 2000). In competitive coevolutionary algorithms, however the increased fitness of one of the species implies a diminution in the fitness of the other species. This evolutionary pressure tends to produce new strategies in the populations involved so as to maintain their chances of survival. This "arms race" ideally increases the capabilities of the species until they reach an optimum. Several methods have been developed to encourage the arms race (Angeline, P.J. & Pollack, J.B., 1993) (Ficici, S.G. & Pollack, J.B., 2001) (Rosin, C. & Belew, R., 1997), but these coevolution methods only consider interaction between species and neglect the effects of the change of the environment on the species. Actually the concept of coevolution is also derived from ecologic science. In ecology, much of the early theoretical work on the interaction between species started with the Lotka-Volterra model of competition (Yuchang, S. & Xiaoming, C., 1996). The model itself was a modification of the logistic model of the growth of a single population and represented the result of competition between species by the change of the population size of each species. Although the model could not embody all the complex relations between species, it is simple and easy to use. So it has been accepted by most of the ecologists.

## 3. Coevolution Based Adaptive Monte Carlo Localization

To overcome the limitations of MCL, samples are clustered into different species. Samples in each species have similar characteristics, and each of the species represents a hypothesis of the place where the robot is located. The Lotka-Volterra model is used to model the competition between species. The competition for limited resources will lead to the extinction of some species, and at the same time make some species become more complex so as to coexist with each other. The environment resources which represent uncertainty of the pose of the robot will change over time, so the total sample size will also change over time. Compared with other coevolution models, our model involves the effects of competition between species as well as that of the change of environment on species.

*3.1. Initial Species Generation*
In the traditional MCL, the initial samples are randomly drawn from a uniform distribution for global localization. If a small sample size is used, few of the initial samples will fall in the regions where the desired posterior density is large, so MCL will fail to localize the robot correctly. In this paper, we propose an efficient initial sample selection method, and at the same time the method will cluster the samples into species.

In order to select the samples that can represent the initial location of the robot well, a large test sample set $S_{test} = \{(\tilde{x}_0^{(1)}, \tilde{w}_0^{(1)}), \cdots, (\tilde{x}_0^{(N_{test})}, \tilde{w}_0^{(N_{test})})\}$ with $N_{test}$ samples is drawn from a uniform distribution over the state space, here $\tilde{w}_0^{(j)} = p(y_0 | \tilde{x}_0^{(j)})$. Then the multi-dimensional state space of the robot is partitioned into small hyper-rectangular grids of equal size. And samples in $S_{test}$ are mapped into the grids. The weight of each grid is the average importance factor of the samples that fall in it. A threshold $T = \mu$ is used to classify the grids into two groups, here the coefficient $\mu \in (0,1)$. Grids with weight larger than $T$ are picked out to form a grid set $V$. The initial sample size $N_0$ is defined by:

$$N_0 = \eta |V| / \overline{w}_0 \qquad (5)$$

Where $\eta$ is a predefined parameter, $\overline{w}_0$ is the average weight of grids in set $V$, and $|V|$ is the number of grids in set $V$. This equation means that if the robot locates in a small area with high likelihood, a small initial sample size is needed.

Using the network defined through neighborhood relations between the grids, the set $V$ is divided into connected regions (i.e. sets of connected grids). Assuming there are $\Omega$ connected regions, these connected regions are used as seeds for the clustering procedure. A city-block distance is used in the network of grids. As in image processing field, the use of distance and seeds permits to define influence zones, and the boundary between influence zones is known as SKIZ (skeleton by influence zone) (Serra, J., 1982). So the robot's state space is partitioned into $\Omega$ parts. And $N_0^{(i)}$ samples which have the largest importance factor will be selected from the test samples falling in the $i$th part.

$$N_0^{(i)} = N_0 \cdot \overline{w}_0(i) / \sum_{k=1}^{\Omega} \overline{w}_0(k) \qquad (6)$$

Where $\overline{w}_0(k)$ is the average weight of grids in the $k$th part of the state space. The selected $N_0^{(i)}$ samples from the $i$th part form an initial species of $N_0^{(i)}$ population size.

*3.2. Inter-Species Competition*

Inspired by ecology, when competing with other species the population growth of a species can be modeled using the Lotka-Volterra competition model. Assuming there are two species, the Lotka-Volterra competition model includes two equations of population growth, one for each of two competing species.

$$\frac{dN_t^{(1)}}{dt} = r^{(1)} N_t^{(1)} (1 - \frac{N_t^{(1)} + \alpha_t^{(12)} N_t^{(2)}}{K_t^{(1)}}) \qquad (7)$$

$$\frac{dN_t^{(2)}}{dt} = r^{(2)} N_t^{(2)} (1 - \frac{N_t^{(2)} + \alpha_t^{(21)} N_t^{(1)}}{K_t^{(2)}}) \qquad (8)$$

Where $r^{(i)}$ is the maximum possible rate of population growth, $N_t^{(i)}$ is the population size and $K_t^{(i)}$ is the upper limit of population size of species $i$ that the environment resource can support at time step $t$ respectively, and $\alpha_t^{(ij)}$ refers to the impact of an individual of species $j$ on population growth of species $i$; here $i, j \in \{1,2\}$. Actually, The Lotka-Volterra model of inter-specific competition also includes the effects of intra-specific competition on population of the species. When $\alpha_t^{(ij)}$ or $N_t^{(i)}$ equals 0, the population of the species $j$ will grow according to the logistic growth model which models the intra competition between individuals in a species.

These equations can be used to predict the outcome of competition over time. To do this, we should determine equilibria, i.e. the condition that population growth of both species will be zero. Let $dN_t^{(1)}/dt = 0$ and $dN_t^{(2)}/dt = 0$. If $r^{(1)}N_t^{(1)}$ and $r^{(2)}N_t^{(2)}$ do not equal 0, we get two line equations which are called the isoclines of the species. They can be plotted in four cases, as are shown in Fig.1. According to the figure, there are four kinds of competition results determined by the relationship between $K_t^{(1)}$, $K_t^{(2)}$, $\alpha_t^{(12)}$ and $\alpha_t^{(21)}$.

(a) When $K_t^{(2)}/\alpha_t^{(21)} < K_t^{(1)}$, $K_t^{(1)}/\alpha_t^{(12)} > K_t^{(2)}$ for all the time steps, species 1 will always win and the balance point is $N_t^{(1)} = K_t^{(1)}, N_t^{(2)} = 0$.

(b) When $K_t^{(2)}/\alpha_t^{(21)} > K_t^{(1)}$, $K_t^{(1)}/\alpha_t^{(12)} < K_t^{(2)}$ for all the time steps, species 2 will always win and the balance point is $N_t^{(1)} = 0, N_t^{(2)} = K_t^{(2)}$.

(c) When $K_t^{(2)}/\alpha_t^{(21)} < K_t^{(1)}, K_t^{(1)}/\alpha_t^{(12)} < K_t^{(2)}$ for all the time steps, they can win each other; the initial population of them determines who will win.

(d) When $K_t^{(2)}/\alpha_t^{(21)} > K_t^{(1)}, K_t^{(1)}/\alpha_t^{(12)} > K_t^{(2)}$ for all the time steps, there is only one balance point and they can coexist with their own population size.

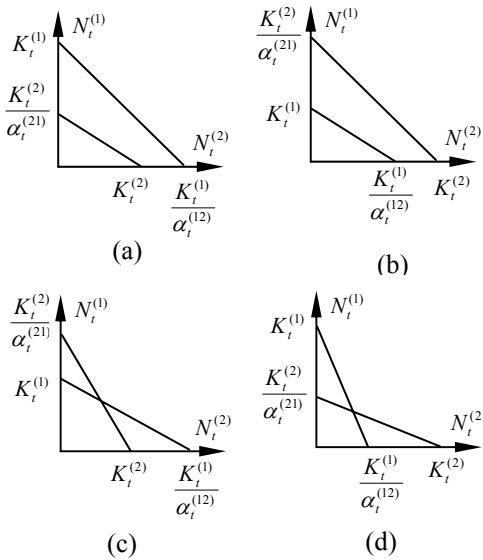

Fig. 1. The isoclines of two coevolution species

For an environment that includes $\Omega$ species, the competition equation can be modified as:

$$\frac{dN_t^{(i)}}{dt} = r^{(i)}N_t^{(i)}(1 - \frac{N_t^{(i)} + \sum_{j=1, j\neq i}^{\Omega} \alpha_t^{(ij)} N_t^{(j)}}{K_t^{(i)}}) \quad (9)$$

### 3.3. Environment Resources

Each species will occupy a part of the state space, which is called living domain of that species. Let matrix $Q_t^{(i)}$ represent the covariance matrix calculated using the individuals in a species $i$. $Q_t^{(i)}$ is a symmetric matrix of $n \times n$, here $n$ is the dimension of the state. Matrix $Q_t^{(i)}$ can be decomposed using singular value decomposition:

$$Q_t^{(i)} = UDU^T \quad (10)$$
$$U = (e_1, \cdots, e_n) \in R^{n \times n} \quad (11)$$
$$D = diag(d_1, \cdots, d_n) \quad (12)$$

Where $d_j$ is the variance of species $i$ in direction $e_j$, $e_j \in R^n$ is a unit vector, and

$$e_j^T \cdot e_k = \begin{cases} 1 & j = k \\ 0 & j \neq k \end{cases} \quad (13)$$

We define the living domain of the species $i$ at time $t$ to be an ellipsoid with radius of $2\sqrt{d_j}$ in direction $e_j$, and it is centered at $\bar{x}_t^{(i)}$ which is the weighted average of the individuals in species $i$. The size of the living domain is decided by:

$$A_t^{(i)} = 2^n C_n \prod_{j=1}^n d_j \quad (14)$$

Where $C_n$ is a constant, depending on $n$, for example $C_2 = \pi$, $C_3 = 4\pi/3$. Actually the living domain of a species reflects the uncertainty of the robot's pose estimated according to that species, and it is similar to the uncertainty ellipsoid (Herbert, M. *et al*, 1997).

Environment resources are proportional to the size of the living domain. The environment resources occupied by a species are defined as:

$$R_t^{(i)} = \begin{cases} \delta \cdot A_t^{(i)} & A_t^{(i)} > \varepsilon \\ \delta \cdot \varepsilon & A_t^{(i)} \leq \varepsilon \end{cases} \quad (15)$$

Where $\delta$ is the number of resources in a unit living domain, and $\varepsilon$ is the minimum living domain a species should maintain. Assuming a species can plunder the resources of other species through competition, i.e. the environment resources are shared by all species. And the number of individuals that a unit of resource can support is different for species with different fitness. The upper limit of population size that the environment resources can support of a species is determined by:

$$K_t^{(i)} = \exp(1 - \overline{w}_t^{(i)}) \cdot R_t \qquad (16)$$

Where parameter $R_t = \sum_{i=1}^{\Omega} R_t^{(i)}$ is the total resources of the system and $\overline{w}_t^{(i)}$ is the average importance factor of species *i*. It is obvious that the environment resources will change over time. In the beginning, the pose of the robot is very uncertain, so the environment resources are abundant. When the pose of robot becomes certain after running for some time, the living domains will become small and the environment resources will also be reduced. The upper limit of population of species will change according to the environment resources, but the change of the resources will not affect the competition results of the species. Supposing $R_t = \lambda R_{t-1}$ and there are two species, upper limit of the population of species will be $K_t^{(1)} = \lambda K_{t-1}^{(1)}$ and $K_t^{(2)} = \lambda K_{t-1}^{(2)}$. This will not change the relation between $K_t^{(2)}/\alpha_t^{(21)}$ and $K_t^{(1)}$, and that between $K_t^{(1)}/\alpha_t^{(12)}$ and $K_t^{(2)}$ in the Lotka-Volterra competition model.

### 3.4. Intra-Species Evolution

Since genetic algorithm and sequential Monte Carlo importance sampling have many common aspects, Higuchi, T. (1997) has merged them together. In CEAMCL the genetic operators, crossover and mutation, are applied to search for optimal samples in each species independently. The intra-species evolution will interact with inter-species competition: the evolution of individuals in a species will increase its ability for inter-species competition, so as to survive for a longer time.

Because the observation density $p(y_t | x_t)$ includes the most recent observed information of the robot, it is defined as the fitness function. The two operators: crossover and mutation, work directly over the floating-points to avoid the trouble brought by binary coding and decoding. The crossover and mutation operator are defined as following:

Crossover: for two parent samples $(x_t^{(p1)}, w_t^{(p1)})$, $(x_t^{(p2)}, w_t^{(p2)})$, the crossover operator mates them by formula (17) to generate two children samples.

$$\begin{cases} x_t^{(c1)} = \xi x_t^{(p1)} + (1-\xi) x_t^{(p2)} \\ x_t^{(c2)} = (1-\xi) x_t^{(p1)} + \xi x_t^{(p2)} \\ w_t^{(c1)} = p(y_t | x_t^{(c1)}) \\ w_t^{(c2)} = p(y_t | x_t^{(c2)}) \end{cases} \qquad (17)$$

Where $\xi \sim U[0,1]$, and $U[0,1]$ represents uniform distribution. And two samples with the largest importance factors are selected from the four samples for the next generation.

Mutation: for a parent sample $(x_t^{(p)}, w_t^{(p)})$, the mutation operator on it is defined by formula (18).

$$\begin{cases} x_t^{(c)} = x_t^{(p)} + \tau \\ w_t^{(c)} = p(y_t | x_t^{(p)}) \end{cases} \qquad (18)$$

Where $\tau \sim N(0, \Sigma)$ is a three-dimensional vector and $N(0, \Sigma)$ represents normal distribution. The sample with larger importance factor is selected from the two samples for next generation. In CEAMCL, the crossover operator will perform with probability $p_c$ and mutation operator will perform with probability $p_m$. Because the genetic operator can search for optimal samples, the sampling process is more efficient and the number of samples required to represent the posterior density can be reduced considerably.

### 3.5. CEAMCL Algorithm

The coevolution model is merged into the MCL, and the new algorithm is termed coevolution based adaptive Monte Carlo localization (CEAMCL). During localization, if two species cover each other and there is no valley of importance factor between them, they will be merged; and a species will be split if grids occupied by the samples can be split into more than one connected regions as in initial species generation. This is called splitting-merging process. The CEAMCL algorithm is described as following:

- Initialization: select initial samples and cluster the samples into $\Omega$ species; for each species *i*, let $dN_0^{(i)}/dt = 0$; and let $t = 1$.

- Sample size determination: if $dN_{t-1}^{(i)}/dt > 0$, draw $dN_{t-1}^{(i)}/dt$ samples randomly from the living domain of the *i*th species and merge the newly drawn samples to $S_{t-1}^{(i)}$; let $N_t^{(i)} = \max(N_{t-1}^{(i)} + dN_{t-1}^{(i)}/dt, 0)$

- Resampling: for each species *i*, resample $N_t^{(i)}$ samples from $S_{t-1}^{(i)}$ according to $w_{t-1}^{(i)}$;

- Importance Sampling: for each species *i*, sample state $x_t^{(ij)}$ from $p(x_t | x_{t-1}^{(ij)}, u_{t-1})$ for each of the $N_t^{(i)}$ possible state $x_{t-1}^{(ij)}$; and evaluate the importance factor $w_t^{(ij)} = p(y_t | x_t^{(ij)})$.

- Intra-species evolution: for each species *i* randomly draw two samples, and mate them with probability $p_c$, repeat this for $N_t^{(i)}/2$ times; then randomly draw one sample from species *i*, and mutate it with probability $p_m$, repeat this for $N_t^{(i)}$ times.

- Splitting-merging process: split and merge the species using the rules defined in the splitting-merging process.

- Calculating the sample size increment: for each species i calculate the upper limit population size of species *i*, and calculate $dN_t^{(i)}/dt$ using equation (9).

- Summary: The species whose average importance factor is the largest is assumed to be the location of the robot, and the importance factors of each species are normalized independently.
- $t=t+1$; go to step *2)* if not stop.

*3,6. Computational Cost*

Compared with MCL, CEAMCL requires more computation for each sample. But the sampling process is more efficient in CEAMCL. So it can considerably reduce the number of samples required to represent the posterior density.

The resampling, importance factor normalization and calculating statistic properties have almost the same computational cost per sample for the two algorithms. We denote the total cost of each sample in these calculations as $T_r$. The importance sampling step involves drawing a $n$ dimensional-vector according to the motion model $p(x_t | x_{t-1}^{(j)}, u_{t-1})$ and computing the importance factor whose computational costs are denoted as $T_s$ and $T_f$ respectively. The additional computational cost for CEAMCL arises from the intra-species evolution step and the splitting-merging process. The evolution step computes the importance factor with probability $p_c$ in crossover and with probability $p_m$ in mutation. The other computation in evolution step includes drawing a $n$ dimensional-vector from a Gaussian distribution, drawing a random number from uniform distribution of $U[0,1]$ and several times of simple addition, multiplication and comparison. Since the other computational cost in evolution step is almost the same as $T_s$ which also includes drawing a $n$ dimensional-vector from a Gaussian distribution, the total computational cost for each sample in evolution step is approximated by $(p_c + p_m)T_f + T_s$. The splitting or merging probability of species in each time step is small, especially when the species become stable no species need to be split or merged, so the computational cost of the splitting-merging process denoted as $T_m$ is small. And the computational costs of other steps in CEAMCL are not related to the sample size, so they can be neglected. Defining $N_M$ and $N_C$ as the number of samples in MCL and CEAMCL respectively, the total computational costs for one of the iteration in localization $T_M$ and $T_C$ are given by:

$$T_M = N_M(T_f + T_s + T_r) \quad (19)$$

$$T_C \approx N_C((1+p) \cdot T_f + 2T_s + T_r + T_m) \quad (20)$$

Where $p = p_c + p_m$. The most computationally intensive procedure in localization is the computation of the importance factor which has to deal with the high dimensional sensor data, so $T_f$ is much larger than the other terms. It is safe to draw the following rule:

$$T_C \approx (1+p) \cdot T_M (N_C / N_M) \quad (21)$$

## 4. Experimental Results

We have evaluated CEAMCL in the context of indoor mobile robot localization using data collected with a PioneerⅡ robot. The data consist of a sequence of laser range-finder scans along with odometry measurements annotated with time-stamps to allow systematic real-time evaluations. The experimental field is a large hall in our laboratory building whose size is of $15 \times 15 \text{ m}^2$, and the hall is partitioned into small rooms using boards to form a highly symmetric map shown in Fig 2.

In the initial species generation, we only use the x-y position and don't use direction to cluster the samples. The map is divided into $150 \times 150$ grids. The test sample size $N_{test}$ equals 1000000; the threshold parameter $\mu$ equals 0.85; and parameter $\eta$ is 2. The initial species generation results are shown in Fig 2. It is obvious that even in the same environment the initial sample sizes are different when the robot is placed in different places. The real position of robot is marked using a small circle.

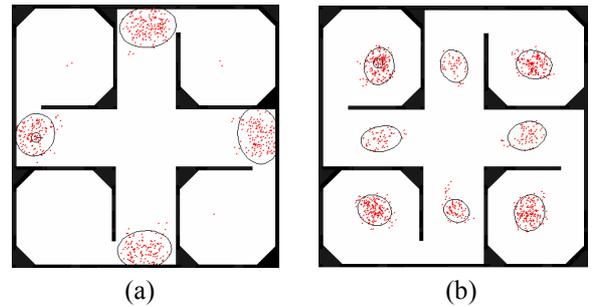

(a) (b)
Fig.2. The initial species for localization. (a) 4 species with 537 samples (b) 8 species with 961 samples

In the localization experiments, the robot was placed at the center of one of the four rooms, and it was commanded to go to a corner of another room. 5 times of experiments were conducted for each center. The three algorithms MCL, GMCL and CEAMCL were applied to localize the robot using the data collected in the experiments. Here GMCL is genetic Monte Carlo localization which merges genetic operators into MCL but without coevolution mechanism. The parameter $\alpha_t^{(ij)} = \overline{w}_t^{(j)} / \overline{w}_t^{(i)}$, where $\overline{w}_t^{(i)}$ is the fitness (average importance factor) of species *i*; parameter $r^{(i)}$, the maximum possible rate of population growth of species, equals 0.2; and parameter $\varepsilon$, the minimum living domain of species, equals 0.5m$^2$; parameter $\delta$, the number of resources in a unit living domain which is 1 m$^2$ in this paper, equals 80; the crossover probability $p_c$ is 0.85; and the mutation probability $p_m$ is 0.15.

The success rate of the three algorithms to track the hypothesis corresponding to the robot's real pose until the robot reaches the goal is shown in table 1. In table 1, the converged time is the time when the samples converged to the four most likely positions; the expired time is the time when one of the hypotheses vanished. The data shows that the CEMCL can converge to the most likely positions as fast as GMCL, but it can maintain the hypotheses for a much longer time than the other two algorithms. This is because the species with much lower fitness will die out because of the competition between species, and the species with similar fitness can coexist with each other for a longer time. Figure 3 shows two inter moments when the robot ran from the center of room 1 to the goal using the CEAMCL algorithm.

|  | CEAMCL | GMCL | MCL |
|---|---|---|---|
| Converged time (s) | 6.87 | 6.41 | 9.76 |
| Expired time (s) | Never | 30.44 | 36.46 |
| Success rate | 100% | 82% | 86% |

Table 1. Success rate of multi-hypotheses tracking

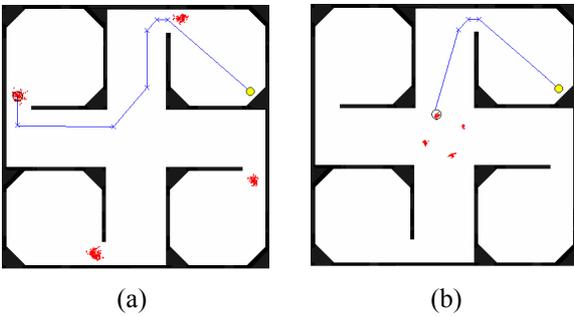

(a)                  (b)
Fig. 3. Localization based on CEAMCL. (a) samples at 7th second (b) samples at 16th second

To compare the localization precision of the three algorithms, we use the robot position tracked by using MCL with 5000 samples in the condition of knowing the initial position to be the reference robot position. The average estimation errors along the running time are shown in Fig 4. Since the summary in CEAMCL is based on the most likely species and the genetic operator in intra-species evolution can drive the samples to the regions with large importance factors, so localization error of CEAMCL is much lower. Although GMCL almost has the same precision as CEAMCL after some time, GMCL is much more likely to produce premature convergence in symmetric environment.

The computational time needed for each iteration with 961 initial samples on a computer with a CPU of PENIUM 800 is shown in Fig 5. Because $p_c + p_m = 1$, the computational time needed for each iteration of CEAMCL is almost twice of that of MCL with the same size of sample set. But since the sample size of CEAMCL is adaptively adjusted during the localization process, the computational time for each iteration of CEAMCL becomes less than that of MCL after some time. From Fig 4. and Fig 5. we can see that the CEAMCL can make precise localization with a small sample size. The changes of the total environment resources and the total number of samples are shown in Fig 6. From the figure we can see that the resources will be reduced when the position becomes certain, the total sample size needed for robot localization will also be reduced.

The parameter $\delta$ which represents the number of resources in a unit living domain is important in CEAMCL, because it will affect the competition between species. Large $\delta$ will reduce the competition between the species since there is enough resource for them. The curve of total sample size with different $\delta$ is shown in Fig 7. The growth rate $r^{(i)}$ is another important parameter. Large $r^{(i)}$ will increase the rate of convergence to the species that have larger fitness. But if $r^{(i)}$ is too large it may cause premature convergence.

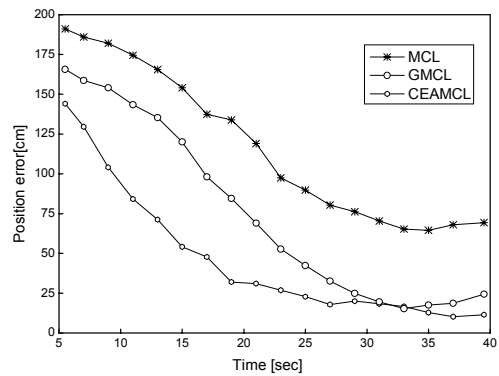

Fig. 4. Estimation Error

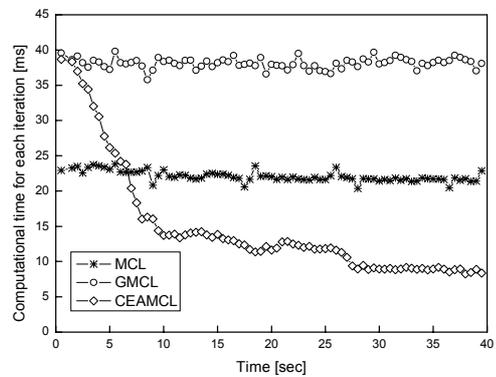

Fig. 5. Computational time for each iteration

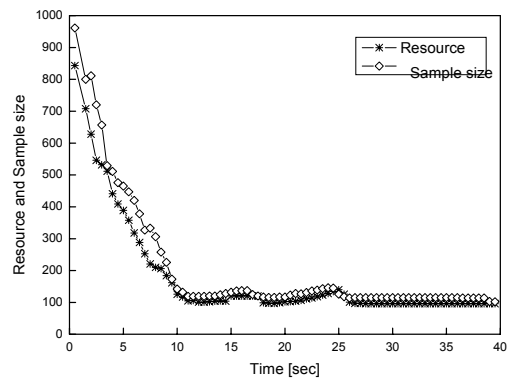

Fig. 6. Change of resource and sample size

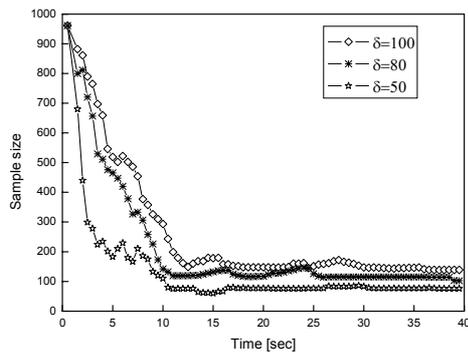

Fig. 7. Effect of $\delta$ on sample size

## 5. Conclusion

An adaptive localization algorithm CEAMCL is proposed in this paper. Using an ecological competition model, CEAMCL can adaptively adjust the sample size according to the total environment resource, which represents uncertainty of the position of the robot. Coevolution between species ensures that the problem of premature convergence when using MCL in highly symmetric environments can be solved. And genetic operators used for intra-species evolution can search for optimal samples in each species, so the samples can represent the desired posterior density better. Experiments prove that CEAMCL has the following advantages: (1) it can adaptively adjust the sample size during localization; (2) it can make stable localization in highly symmetric environment; (3) it can make precise localization with a small sample size.